\documentclass[sigconf]{acmart}

\usepackage{hyperref}
\usepackage{arydshln} 
\usepackage{amsmath}
\usepackage{amsfonts}
\usepackage{bm}
\usepackage{multirow} 
\usepackage{graphicx}
\usepackage{subcaption}
\usepackage{booktabs}
\usepackage{bbding}
\usepackage{enumitem}
\newcommand{\etal}{\textit{et al.}}
\newcommand{\eg}{\textit{e.g.}}
\newcommand{\ie}{\textit{i.e.}}
\usepackage{pifont}
\newcommand{\cmark}{\ding{51}}%
\newcommand{\xmark}{\ding{55}}%

\AtBeginDocument{%
  }

\copyrightyear{2025}
\acmYear{2025}
\setcopyright{acmlicensed}\acmConference[ICMR '25]{Proceedings of the 2025 International Conference on Multimedia Retrieval}{June 30-July 3, 2025}{Chicago, IL, USA}
\acmBooktitle{Proceedings of the 2025 International Conference on Multimedia Retrieval (ICMR '25), June 30-July 3, 2025, Chicago, IL, USA}
\acmDOI{10.1145/3731715.3733442}
\acmISBN{979-8-4007-1877-9/2025/06}

\begin{document}

\title{TeDA: Boosting Vision-Lanuage Models for Zero-Shot 3D Object Retrieval via Testing-time Distribution Alignment}

\author{Zhichuan Wang}
\affiliation{%
  \institution{Huazhong Agricultural University}
  \city{Wuhan}
  \state{Hubei}
  \country{China}
}
\email{wzc_65@webmail.hzau.edu.cn}

\author{Yang Zhou}
\affiliation{%
  \institution{Shenzhen University}
  \city{Shenzhen}
  \state{Guangdong}
  \country{China}}
  \email{zhouyangvcc@szu.edu.cn}

\author{Jinhai Xiang}
\affiliation{%
 \institution{Huazhong Agricultural University}
  \city{Wuhan}
  \state{Hubei}
  \country{China}
  }
\email{jimmy_xiang@mail.hzau.edu.cn}

\author{Yulong Wang}
\authornote{Corresponding author}
\affiliation{%
 \institution{Huazhong Agricultural University}
  \city{Wuhan}
  \state{Hubei}
  \country{China}
  }
\email{wangyulong6251@gmail.com}

\author{Xinwei He}
\authornotemark[1]
\affiliation{%
  \institution{Huazhong Agricultural University}
  \city{Wuhan}
  \state{Hubei}
  \country{China}}
  \email{xwhe@hzau.edu.cn}

\renewcommand{\shortauthors}{Zhichuan Wang et al.}

\begin{abstract}
Learning discriminative 3D representations that generalize well to unknown testing categories is an emerging requirement for many real-world 3D applications.
Existing well-established methods often struggle to attain this goal due to insufficient 3D training data from broader concepts. 
Meanwhile, pre-trained large vision-language models (e.g., CLIP) have shown remarkable zero-shot generalization capabilities. Yet, they are limited in extracting suitable 3D representations due to substantial gaps between their 2D training and 3D testing distributions.
To address these challenges, we propose \textbf{Te}sting-time \textbf{D}istribution \textbf{A}lignment (\textbf{TeDA}), a novel framework that adapts a pretrained 2D vision-language model CLIP for unknown 3D object retrieval at test time. To our knowledge, it is \textbf{the first work} that studies the test-time adaptation of a vision-language model for 3D feature learning.
TeDA projects 3D objects into multi-view images, extracts features using CLIP, and refines 3D query embeddings with an iterative optimization strategy by confident query-target sample pairs in a self-boosting manner. 
Additionally, TeDA integrates textual descriptions generated by a multimodal language model (InternVL) to enhance 3D object understanding, leveraging CLIP's aligned feature space to fuse visual and textual cues. Extensive experiments on four open-set 3D object retrieval benchmarks demonstrate that TeDA greatly outperforms state-of-the-art methods, even those requiring extensive training. We also experimented with depth maps on Objaverse-LVIS, further validating its effectiveness. Code is available at \href{https://github.com/wangzhichuan123/TeDA}{https://github.com/wangzhichuan123/TeDA}.
\end{abstract}



\maketitle

\section{Introduction}
\label{sec:intro}

In recent years, we have witnessed the proliferation of 3D models/objects created from diverse fields, such as gaming, virtual reality and 3D printing.
It poses a critical challenge for efficient 3D data management, involving accurate search, retrieval and analysis of 3D objects. 
For this, existing works~\cite{su2015multi, qi2017pointnet,zhao2019pointweb, maturana2015voxnet} typically learn 3D representations on various data formats (\emph{e.g.}, voxels, multi-view images, and point clouds), with the help of large-scale labeled 3D benchmarks, such as ShapeNet~\cite{chang2015shapenet}. 
Despite notable advancements, these methods typically assume the same training and testing distributions. However, in real-world scenarios, 3D objects of new unknown categories are continually emerging at testing time, making it hard for those well-trained models to extract suitable generalized representations for 3D analysis. 
Consequently, there is an urgent need for scalable methods that can rapidly adapt to unknown testing categories in real-world environments.

Recent research has explored two primary paradigms to enhance unknown category generalization. 
One is the open-set paradigm~\cite{feng2023hypergraph,feng2022shrec}, which focuses on learning generalized representations from the training set of known categories directly. HGM${^2}$R~\cite{feng2023hypergraph} employs hypergraphs to model the relationships between seen and unseen categories. 
Yet, it depends on multi-modal inputs and availability of test data. 
The other is the zero-shot paradigm, which leverages large 2D foundation models (e.g., CLIP) for multi-view images or 3D ones (OpenShape~\cite{liu2024openshape}, ULIP-2~\cite{xue2024ulip}, and Uni3D~\cite{zhou2023uni3d}) for point clouds. Benefiting from massive data pertaining, they display robust zero-shot generalization capabilities to unseen categories. However, they require large-scale 3D datasets with broader concepts for resource-intensive training. 
Recently, Test-Time Adaptation (TTA) has emerged as a promising approach to mitigate the distribution drift in many 2D vision tasks~\cite{karmanov2024efficient,feng2023diverse,shu2022test,sun2020test,qian2024intra}. By adapting models to test data during testing, it efficiently improves performance on downstream tasks. 
Inspired by its success, we ask: can we adapt extracted 3D embeddings from off-the-shelf 2D foundation models like CLIP swiftly during testing?

\begin{figure}[h]
\centering
\includegraphics[width=0.95\linewidth]{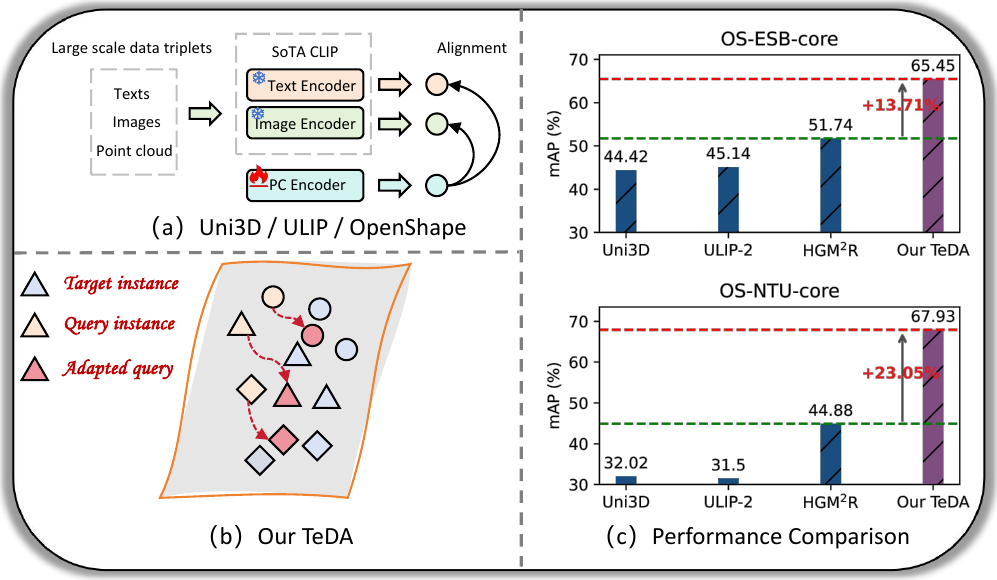}
\caption{(a) Existing methods like Uni3D train a point cloud encoder by aligning visual and textual encoders from CLIP using large-scale data triplets, which requires substantial time and GPU resources. In contrast, (b) our method leverages rapid test-time adaptation to narrow the gap between query and target instances, without the need for training or model adjustments, preserving the inherent priors and generalization capabilities of CLIP. As shown in (c), our method achieves a significant improvement in retrieval performance.}
\label{fig:perf_com}
\end{figure}

However, it is nontrivial to apply current TTA methods to 3D retrieval involving both query and target sets instead of only one set, especially when testing categories are completely unknown. 
For this, we propose a non-parametric iterative methodology named \textbf{TeDA} to efficiently perform \textbf{Te}sting-time \textbf{D}istribution \textbf{A}lignment for multi-view embeddings from a pretrained CLIP model. 
The main idea behind \textbf{TeDA} is to select high-confidence query-target sample pairs to learn how to adjust query features in the direction of the target features in a self-boosting way. 
Our approach starts by projecting each 3D object from both query and target sets into 2D multi-view images. Next, we extract and aggregate view features from a pretrained CLIP model with mean-pooling, which serves as the initial 3D descriptors before adaptation. 
For adaptation, we first calculate the pairwise instance-level similarity matrices among 3D descriptors of unknown categories from both sets at test time. Based on it, we select sample pairs with higher similarities to further \emph{iteratively} refine the query features towards the target distributions.
By aligning the two feature distributions with standard Kullback-Leibler (KL) divergence, we effectively and efficiently bridge the domain gaps between CLIP's pretrained data and multi-view projections of 3D objects. 

Moreover, to better harness the ability of CLIP as a dual-stream network, we further propose to incorporate additional textual knowledge from off-the-shelf multi-modal language models to describe 3D objects comprehensively. Specifically, we employ InternVL~\cite{chen2024internvl} to generate a descriptive sentence for each 3D object. 
The motivation behind this is straightforward: text is a more stable modality, and a well-crafted description of an object can be highly beneficial for distinguishing unseen categories. 
Thanks to CLIP's aligned feature space for image and text inputs, we can easily embed the acquired descriptions via CLIP encoder and fuse the visual and textual cues in CLIP space with simple operations, such as addition, to enhance retrieval features effectively. We comprehensively evaluate TeDA on four popular open-set 3DOR benchmarks~\cite{feng2023hypergraph}, surpassing state-of-the-arts by significant margins, demonstrating its remarkable potential in tackling the unknown category issue in 3DOR. 

In summary, the main contributions are the following:
\begin{itemize}
 \item To our knowledge, this is the first work that attempts to adapt a 2D pretrained vision-language model for unknown 3D object retrieval \emph{at test time}. We propose a multi-view-based framework, named TeDA, which refines pseudo labels based on 3D data distributions in a training-free manner.
 \item We propose to integrate textual information from a large multi-modal language model (InternVL) as extra cues to enrich the 3D object understanding. By combining both textual and visual embeddings, we derive more discriminative and generalizable 3D features for adaptation. 
 \item Extensive experiments demonstrate that TeDA achieves superior performance. As a training-free approach, it even surpasses methods that require extensive training. 
\end{itemize}

\section{Related Work}
\label{sec:related_work}

\noindent\textbf{3D Object Retrieval (3DOR).}
In general, we can group existing works into two classes: closed-set and open-set 3D object retrieval.  

\textbf{1) Closed-set 3DOR.} They assume a closed-set setting where train and testing sets contain 3D objects from the same label set. Modern works generally learn 3D representations directly with deep neural networks for embedding. Overall, these methods can be coarsely divided into two types: \emph{Model-based} and \emph{View-based}. 
\emph{Model-based} methods deals with raw 3D formats, including voxel grids~\cite{maturana2015voxnet,wu20153d,wang2017cnn}, point clouds~\cite{qi2017pointnet, qi2017pointnet++, wang2019dynamic, liu2019relation, guo2020deep,cheng2021net,you2018pvnet}, or meshes~\cite{feng2019meshnet,liang2022meshmae}. These methods have the merit of capturing rich 3D geometry structures. However, the 3D formats often induce computation challenges, \eg, cubic computation complexity for voxels and the irregular and unordered issue of point clouds and meshes.
On the other hand, \emph{view-based} methods~\cite{kanezaki2018rotationnet,han20193d2seqviews, esteves2019equivariant,feng2018gvcnn,su2015multi,he2018triplet,he2019view,he2020improved} project 3D objects into a set of 2D images and then use a convolutional neural network to embed each 2D image. Finally, a compact descriptor is aggregated. 
Compared with 3D formats, 2D view images are regular grids and seamlessly fit many off-the-shelf 2D backbones. 
\emph{Despite great progress, they neglect the open nature in real scenarios where 3D objects of unseen test categories emerge frequently.}

\textbf{2) Open-set 3DOR}. Recently, researchers~\cite{feng2023hypergraph,linok2024beyond,feng2022shrec,xu2024semi,xu2024triadic,xu2024assembly,xu2024structure} shifted their attention to the new open-set setup, which aims to retrieve 3D objects of unknown categories.
A specific track SHREC’22~\cite{feng2022shrec} is organized, which suggests that existing 3DOR works easily overfit training categories and do not generalize well to unknown ones. Feng~\etal~\cite{feng2023hypergraph} further proposes using hypergraphs to connect known training and unseen testing categories, based on multi-modal embeddings of 3D objects. \emph{Our work also tackles this setup but only relies on multi-view images. 
More importantly, TeDA can generalize to unseen categories by adapting a pretrained vision-language model at test time, without requiring a 3D training set.} 
It not only simplifies the input requirements but also makes our method more scalable and practical for real-world applications.

\noindent\textbf{Transfering Vision-Language Models for 3D.}
In recent years, there has been a growing interest in applying large vision-language models (VLMs) such as CLIP~\cite{radford2021learning} to 3D representation learning.  
One way is to directly apply them to view/depth images of 3D objects in a zero-shot manner, such as MV-CLIP~\cite{song2023mv}, PointCLIP~\cite{zhang2022pointclip} and PointCLIP V2~\cite{zhu2023pointclip}. Another way is to distill the embedded knowledge to point cloud backbones, such as ULIP~\cite{xue2023ulip} and ULIP-2~\cite{xue2024ulip}. 
However, zero-shot utilization gives suboptimal performance due to domain gaps, while the distillation methods often necessitate curating large 3D datasets of triplet <image, text, point cloud>. 
In this paper, we start with a pretrained vision and language model, \ie, CLIP, by adapting it to downstream 3D distributions at test time. 
To our knowledge, we are the first to explore test time adaptation of VLMs for 3D representation learning and propose a high-performing alternative for learning generalized 3D embeddings. 

\noindent\textbf{Test-time Adaptation.}
Test-time adaptation refers to adapting models to testing data that may have distributional differences from the training data. It is particularly well-suited for real-world applications where models need to be applied across multiple scenarios. Recent works~\cite{karmanov2024efficient,feng2023diverse,shu2022test,zhang2022memo,zhang2021adaptive,wang2020tent,varsavsky2020test,sun2020test} have made significant progress by applying test-time adaptation to 2D multi-task classification, but current methods still rely on fixed category information to generate pseudo-labels, which makes them unsuitable for direct application to 3D open-set retrieval tasks.
To address this limitation, TeDA treats each query sample as a separate instance, constructing query-target pairs to compute pseudo-labels for test-time adaptation, successfully applying test-time adaptation to retrieval domain.

\begin{figure*}[ht]
\centering
\includegraphics[width=\linewidth]{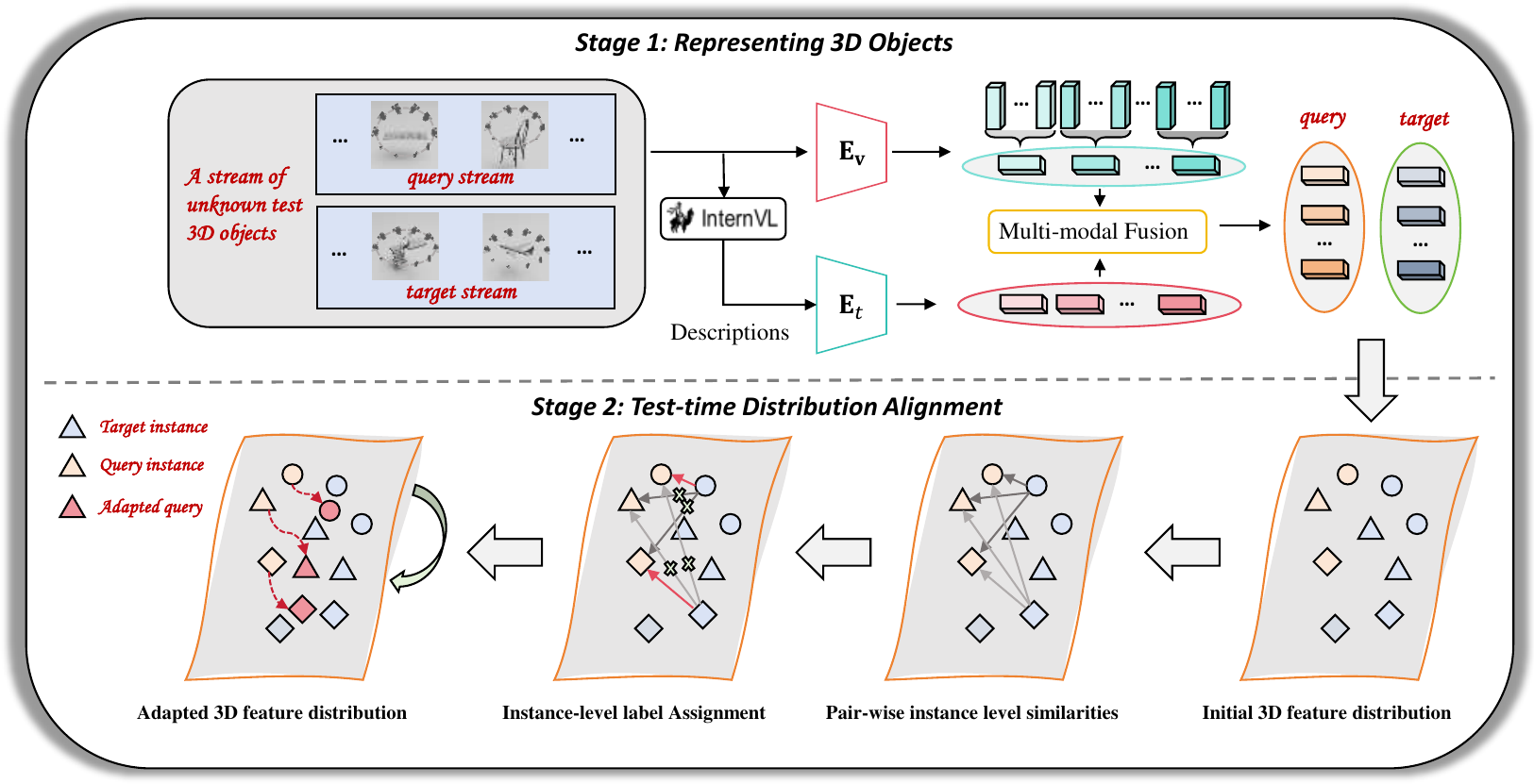}
\caption{Overview of our TeDA. It can be decomposed into two stages, with the first stage representing the query and target streams of 3D objects with off-the-shelf pretraind VLMs (\ie, CLIP and InternVL) while the other one applying test-time distribution alignment for more discriminative 3D descriptors.}
\label{fig:framework}
\end{figure*}

\section{Method}
\label{sec:method}

\subsection{Preliminaries}

\noindent\textbf{Problem definition.}
We aim to adapt 3D embeddings from pre-trained large foundation models to unknown categories at test time, without any retraining or fine-tuning. It is especially advantageous for real-world applications that require rapid adaptation to diverse retrieval scenarios.
Formally, let $\mathcal{Q}$ and $\mathcal{T}$ denote the query and target set, respectively, our goal is to align their embeddings to bridge the distribution gap, while all classes of 3D objects remain hidden during adaption. After adaption, we evaluate the retrieval performance. This task is related to open-set 3D object retrieval, where models are trained on seen categories but tested on unseen categories. It is also related to test-time adaption, though our focus is on the retrieval scenarios which involves mitigating the distribution gaps instead of one. Besides, we do not assume prior knowledge of the category set during adaptation, making the problem more challenging and better suited for real-world applications.

\noindent\textbf{A revisit to CLIP.} CLIP~\cite{radford2021learning} is a vision-language model that performs a proxy task of predicting the correct pairings of text and image. For example, in a classification task, CLIP calculates the corresponding image features and text features using an image encoder $E_\text{visual}(\cdot)$ and a text encoder $E_{\text{text}}(\cdot)$, respectively. The goal of CLIP is to retrieve the most relevant image-text feature pairs in a zero-shot setting, where it can directly match unseen images to textual descriptions without requiring any task-specific training.

\noindent\textbf{A revisit to InternVL.} InternVL~\cite{chen2024internvl} is a multi-modal large language model that not only supports pure text-based question answering but also enables Visual Question Answering (VQA). By providing images along with an appropriate prompt, InternVL generates an initial description for the images.

\subsection{Representing 3D Objects}
In this paper we \emph{only} leverage multi-view projections of 3D objects, as 2D view images can be seamlessly processed by existing 2D foundation models. To comprehensively characterize 3D objects of unknown categories, we employ a combination of visual and textual features. Visual features are adept at capturing the geometric and structural nuances of the objects, while textual features supply semantic context for unknown categories. This synergy results in a more comprehensive and robust representation, facilitating effective adaptation and retrieval. 
The whole 3D feature extraction process can be divided into three steps: 

\noindent\textbf{Step 1: Multi-view Projection.}
To enable CLIP to handle 3D objects, we first project the 3D objects into multi-view images. Specifically, we follow the projection strategy of HGM${^2}$R~\cite{feng2023hypergraph}, where virtual cameras are evenly placed around each 3D object $\mathcal{O}$, generating grayscale images from different angles. These images are represented as a set of views $\{I_{m} \in \mathbb{R}^{1 \times H \times W}\}_{m=1}^M$, where $H$ and $W$ represent the height and width of the view image, respectively, and \( M \) denotes the number of different views captured from various angles.
These multi-view images from different angles can provide a comprehensive representation of a 3D object.

\noindent\textbf{Step 2: Generating Caption.}
To incorporate the text modality during retrieval, we utilize InternVL~\cite{chen2024internvl} to generate an initial description of the object. Specifically, we input the generated multi-view images together with the prompt: "There are images of an object from different angles. Describe this object in one sentence." into InternVL, which produces a semantic description \( q \) of the object. The text modality provides a more stable representation, which is crucial for the open-set 3DOR task, where the model needs to output features that are not only stable but also exhibit strong generalization capabilities. Subsequent experiments have demonstrated that introducing the text modality significantly aids the model in recognizing unseen categories.

\noindent\textbf{Step 3: Fusing Text and Multi-view Features.} 
After obtaining the semantic description $q$ of the 3D object, the next step is to fuse the features from the two modalities. Specifically, we input the multi-view images $\{I_m\}_{m=1}^M$ and the semantic description $sent$ into the CLIP Image Encoder and Text Encoder, respectively, to obtain the corresponding features $\{\mathbf{g}_m \in \mathbb{R}^d\}_{m=1}^M$ and $\mathbf{f}_{\text{text}} \in \mathbb{R}^d$:
\begin{equation}
[\mathbf{g}_1;\mathbf{g}_2;...;\mathbf{g}_M] =  E_{\text{visual}} ([I_1; I_2; ...; I_M]) )  
\end{equation}
\begin{equation}
\mathbf{f}_{\text{text}} = E_{\text{text}}(q)
\end{equation}
where $[;]$ represents concatenation along batch dimension. 
We then utilize \text{mean-pooling}($\cdot$) to average the $M$ view features into global 3D representations $\mathbf{g} \in \mathbb{R}^d$: 
\begin{equation}
    \mathbf{g} = \text{mean-pooling}([\mathbf{g}_1;\mathbf{g}_2;...;\mathbf{g}_M])
\end{equation}

Since CLIP has already aligned the features from both modalities, we can simply add the two modality features together to obtain a fused feature representation. 
The fusion is done as follows:
\begin{equation}
\mathbf{h} = \tanh(\mathbf{g} + \lambda \mathbf{f}_{\text{text}} )
\end{equation}
where \( \lambda \in [0, 1] \) balances the visual and textual features. \( \tanh \) is applied for normalization after the fusion of \( \mathbf{g} \) and \( \mathbf{f}_{t} \). Finally, \( \mathbf{h} \in \mathbb{R}^d \) represents the final descriptor, which forms the basis for subsequent adaptation processes.

\subsection{Bridging Distribution Gaps for Unseen Categories} 
Since CLIP is trained on natural images, its ability to process multi-view images and generalize them to unseen categories is limited. 
To address this, we draw inspiration from~\cite{sohn2020fixmatch,qian2024intra,wu2018unsupervised} and develop a methodology named TeDA to improve the model's instance discrimination at test time through an unsupervised learning paradigm. 

Specifically, we treat each query as a unique instance class and aim to minimize the distribution gap between queries and targets, based on the confident instance-level relationships.
Let $\mathbf{Q} \in \mathbb{R}^{S \times d}, \mathbf{X} \in \mathbb{R}^{N \times d}, \overline{\mathbf{Q}} \in \mathbb{R}^{S \times d}$, denote initial query features,  target features, and optimized query features, respectively, with each row of them representing descriptor vectors for 3D objects. 
At the beginning of adaption, we initialize $\overline{\mathbf{Q}}$ with $\mathbf{Q}$.
Based on them, we calculate the mapping of target feature matrix $\mathbf{X}$ over query features $\mathbf{Q}$ and $\overline{\mathbf{Q}}$, respectively: 
\begin{equation}
\mathbf{P}' = \text{Softmax}(\mathbf{X}\mathbf{Q}^T/\tau_t);
\mathbf{P} = \text{Softmax}(\mathbf{X}\overline{\mathbf{Q}}^T/\tau_i)
\end{equation}
where $\mathbf{P}', \mathbf{P} \in \mathbb{R}^{N \times S}$, each row of which can be viewed as the distribution of a target sample estimated by the original $\mathbf{Q}$ and the learnable proxy $\overline{\mathbf{Q}}$, respectively.

\noindent\textbf{Unsupervised Pseudo-Label Acquisition.} 
In this section, we aim to acquire high-confidence one-hot pseudo-labels, which are generated from soft labels through softmax and thresholding. These pseudo-labels are then used to guide the subsequent optimization process, enabling the model to make confident predictions even for unseen categories. High-confidence pseudo-labels not only eliminate the interference from irrelevant categories but also help in defining the decision boundaries of the classifier. In the absence of fully labeled data, one-hot pseudo-labels provide a powerful supervisory signal similar to that of supervised learning.

In this process, soft labels with high confidence are converted into one-hot labels to eliminate the influence of irrelevant classes. Specifically, for each target sample, the highest predicted probability in the row is compared with a threshold $\alpha$. If the maximum probability exceeds the threshold, the corresponding target distribution is converted to a one-hot encoding by setting the index of the maximum value to 1, while all other positions are set to 0:
\begin{equation}
\small 
\mathbf{P}'[i, j] = 
\begin{cases} 
1, & \text{if } \max (\mathbf{P}'[i, :]) > \alpha \textbf{ \& } j = \arg\max (\mathbf{P}'[i, :]) \\ 
\mathbf{P}'[i, j], & \text{if } \max (\mathbf{P}'[i, :]) \leq \alpha \\
0, & \text{otherwise}
\end{cases}
\end{equation}
where \( i \) is the sample index, and \( j \) is the class index. The symbol `\&' represents the logical "and" between the two conditions. This pseudo-label distribution \( \mathbf{P'} \) can be viewed as a refined version of a one-hot label, focusing only on the relevant class while ignoring the noise from irrelevant ones.

\noindent\textbf{Learning Proxy Modality \( \mathbf{\overline{Q}} \).} 
Subsequently, we optimize the original retrieval distribution based on the obtained pseudo-label distribution \( \mathbf{P'} \). This process is equivalent to minimizing the Kullback-Leibler (KL) divergence between the retrieval probability distribution \( \mathbf{P} \) and the pseudo-label distribution \( \mathbf{P'} \). It is also equivalent to updating the predicted probability distribution by optimizing \( \mathbf{Q} \). 
The KL divergence is computed as:
\begin{equation}
\min_{\overline{\mathbf{Q}}} L(\mathbf{P}', \mathbf{P}(\overline{\mathbf{Q}})) = \sum_i D_{\text{KL}}(\mathbf{P}'_i \parallel \mathbf{P}_i)
\end{equation}

The objective of the optimization is to minimize this divergence, ensuring that the updated retrieval distribution \( \mathbf{P} \) aligns as closely as possible with the pseudo-label distribution \( \mathbf{P'} \). The gradient of this divergence with respect to the query features \( \mathbf{Q} \) is computed, and the query features are updated iteratively using gradient descent:
\begin{equation}
\Delta_{\overline{\mathbf{Q}}} = \mathbf{X}^{T} \left( \frac{\mathbf{X} \cdot \overline{\mathbf{Q}}^{T}}{\tau_i} - \mathbf{P}' \right)
\end{equation}
\begin{equation}
\overline{\mathbf{Q}}^{T} \leftarrow \overline{\mathbf{Q}}^{T} - \frac{\eta}{n_t \cdot \tau_i} \cdot \Delta_{\overline{\mathbf{Q}}}
\end{equation}
where \( \eta \) is the learning rate, \( n_t \) is the number of target instances, and \( \tau_i \) is the temperature scaling factor. This update rule iteratively refines the query feature representation by minimizing the KL divergence, thereby enhancing the retrieval performance.

\noindent\textbf{Final Retrieval Distribution Calculation.}
The retrieval distribution is ultimately derived by performing matrix multiplication between the target features and the updated query features. This is expressed mathematically as:
\begin{equation}
\mathbf{R} =  \overline{\mathbf{Q}} \cdot \mathbf{X}^{T}
\end{equation}
where $\mathbf{R}$ denotes the improved similarity score matrices between the query and the target for evaluation.  

\section{Experiments}
\subsection{Experimental Setup}
\noindent\textbf{Datasets and Evaluation Metrics.}
We evaluate TeDA on four open-set 3D object retrieval benchmarks~\cite{feng2023hypergraph}, including OS-ESB-core, OS-NTU-core, OS-MN40-core, and OS-ABO-core. \emph{OS-ESB-core} has 120 query samples and 452 target samples from 24 unseen categories. \emph{OS-NTU-core} includes 54 unseen categories with 270 query samples and 1271 target samples. \emph{OS-MN40-core} covers 160 probe objects and 9,329 target objects from 32 unseen categories. \emph{OS-ABO-core} consists of 17 categories with 85 samples for the query set and the remaining samples for the target set. 
Based on these datasets, we perform a test-time adaption on the pretrained embeddings of the query and target sets, and report the retrieval performance. 
For evaluation, we use mean Average Precision (mAP), Normalized Discounted Cumulative Gain (NDCG), and Average Normalized Modified Retrieval Rank (ANMRR) metrics.
Precision-Recall curves are also plotted for comprehensive evaluations.

\noindent\textbf{Implementation Details.}
For experiments, we use the same rendering scheme as HGM$^2$R~\cite{feng2023hypergraph}
and project 24 view images of size $256 \times 256$ for each 3D object. 
We employ the pretrained CLIP models~\cite{radford2021learning} with ViT-L/14 as the backbone, as well as the OpenCLIP~\cite{ilharco_gabriel_2021_5143773} version of ViT-L/14. 
To obtain descriptions of 3D object appearances and structures, we use the large generative model InternVL-4B~\cite{chen2024internvl}. In the optimization process, we set the pseudo-label threshold $\alpha$ to 0.6. 
We optimize the embeddings with standard projected gradient descent for 2000 iterations. The initial learning rate is set to 10 and is halved whenever the gradient norm increases. 
All experiments are performed on a single NVIDIA GeForce RTX 4090 GPU.

\subsection{Results on Open-set 3DOR Datasets}

\begin{table*}[h]
\centering
\small 
\setlength{\tabcolsep}{1.0pt}
\caption{Performance Comparison (\%) with state-of-the-art methods on four public open-set 3DOR benchmarks. 
\textbf{Bold} and \underline{underline} indicate the best and second best results, respectively. $^{\ddagger}$ means we provide \emph{ground-truth} category sets for the view selection process in MV-CLIP, which is impractical, as category information for unseen classes is unknown in open-set scenarios.
}
\resizebox{0.95\linewidth}{!}{
\begin{tabular}
{l@{\hspace{1pt}}cccc@{\hspace{10pt}}ccc@{\hspace{10pt}}ccc@{\hspace{10pt}}ccc}
\toprule
\multirow{2}[2]{*}{\textbf{Method}} & \multirow{2}[1]{*}{\textbf{Backbone}} & \multicolumn{3}{c}{\textbf{OS-ESB-core}} & \multicolumn{3}{c}{\textbf{OS-NTU-core}} & \multicolumn{3}{c}{\textbf{OS-MN40-core}} & \multicolumn{3}{c}{\textbf{OS-ABO-core}} \\
\cmidrule(lr){3-5} \cmidrule(lr){6-8} \cmidrule(lr){9-11} \cmidrule(lr){12-14}
 & & \footnotesize{mAP$\uparrow$} & \footnotesize{NDCG$\uparrow$} & \footnotesize{ANMRR$\downarrow$} & \footnotesize{mAP$\uparrow$} & \footnotesize{NDCG$\uparrow$} & \footnotesize{ANMRR$\downarrow$} & \footnotesize{mAP$\uparrow$} & \footnotesize{NDCG$\uparrow$} & \footnotesize{ANMRR$\downarrow$} & \footnotesize{mAP$\uparrow$} & \footnotesize{NDCG$\uparrow$} & \footnotesize{ANMRR$\downarrow$} \\
\midrule

\multicolumn{14}{c}{\textbf{\textit{Training-required}}} \\ 
TCL (CVPR'18)~\cite{he2018triplet}  & \multirow{9}[1]{1.6cm}{\footnotesize{ResNet18, PointNet, 3DShapenets}} 
& 49.31 & 21.89 & 52.68 & 39.37 & 21.23 & 61.00 & 48.11 & 63.83 & 52.30 & 49.33 & 53.86 & 51.05 \\
SDML (SIGIR'19)~\cite{hu2019scalable} &
& 49.59 & 21.75 & 52.36 & 40.16 & 21.52 & 60.49 & 50.75 & 65.70 & 50.22 & 47.44 & 52.79 & 52.42 \\
CMCL (CVPR'21)~\cite{jing2021cross} &
& 50.01 & 21.97 & 53.06 & 41.08 & 21.72 & 59.43 & 51.38 & 65.98 & 49.75 & 49.83 & 50.89 & 50.24 \\
MMSAE (Neurocomputing'19)~\cite{wu2019multi} &
& 49.88 & 22.06 & 53.69 & 40.85 & 21.70 & 59.99 & 52.08 & 66.57 & 49.00 & 50.51 & 53.80 & 50.49 \\
MCWSA (TMM'22)~\cite{zheng2022multi} &
& 49.48 & 21.34 & 53.75 & 39.22 & 20.69 & 62.14 & 48.78 & 63.85 & 51.95 & 45.61 & 51.05 & 54.70 \\
PROSER (CVPR'21)~\cite{zhou2021learning} &
& 48.69 & 21.13 & 53.95 & 39.47 & 21.24 & 60.96 & 49.00 & 64.54 & 51.66 & 50.33 & 53.27 & 50.34 \\
InfoNCE (arXiv'18)~\cite{oord2018representation}  &
& 50.26 & 21.91 & 52.63 & 40.03 & 21.19 & 61.09 & 47.37 & 63.31 & 53.02 & 46.83 & 52.14 & 53.50 \\
HGM$^{2}$R (TPAMI'23)~\cite{feng2023hypergraph} & 
& 51.74 & 22.73 & 51.28 & 44.88 & 22.81 & 56.67 & 64.20 & 72.91 & 38.27 & 63.39 & 57.96 & 37.96 \\

\midrule 

\multicolumn{14}{c}{\textbf{\textit{Training-free}}} \\ 
ULIP (CVPR'23)~\cite{xue2023ulip} & \footnotesize{PointMLP-SLIP} & 33.81 & 16.71 & 68.18 & 20.76 & 13.46 & 78.26 & 24.56 & 38.23 & 72.41 & 34.00 & 39.79 & 64.94 \\
ULIP-2 (CVPR'24)~\cite{xue2024ulip} & \footnotesize{PointBERT-CLIP ViT-G/14} & 45.14 & 21.00 & 59.15 & 31.50 & 17.89 & 68.99 & 32.76 & 48.92 & 65.22 & 44.26 & 49.04 & 55.64 \\
Uin3D (ICLR'24)~\cite{zhou2023uni3d} & \footnotesize{Uni3D-Giant} & 44.42 & 20.92 & 59.96 & 32.02 & 18.04 & 68.49 & 33.21 & 50.51 & 65.11 & 45.92 & 49.79 & 53.96 \\
OpenShape (NeurIPS'23)~\cite{liu2024openshape} & \footnotesize{PointBERT-CLIP ViT-L/14} & 38.58 & 18.81 & 65.10 & 24.71 & 15.02 & 75.18 & 29.64 & 44.79 & 67.64 & 38.65 & 45.56 & 60.90 \\
MV-CLIP$^{\ddagger}$ (arXiv'23)~\cite{song2023mv} & \footnotesize{CLIP ViT-L/14} & 49.81 & 22.75 & 53.71 & 57.71 & 26.46 & 45.25 & 63.74 & 74.85 & 37.80 & 63.07 & 59.18 & 38.67\\
MV-CLIP$^{\ddagger}$ (arXiv'23)~\cite{song2023mv} & \footnotesize{OpenCLIP ViT-L/14} & 58.11 & 24.29 & 44.49 & 57.23 & 26.25 & 45.00 & 60.66 & 72.76 & 40.68 & 59.74 & 57.11 & 41.88 \\
 \textbf{Ours} &  \footnotesize{CLIP ViT-L/14} &  \underline{61.40} &  \underline{25.30} &  \underline{43.73} &  \underline{64.49} &  \underline{28.45} &  \underline{39.62} &  \underline{66.31} &  \underline{74.98} &  \underline{36.22} &  \underline{69.91} &  \underline{59.30} &  \underline{32.41} \\
\textbf{Ours} &  \footnotesize{OpenCLIP ViT-L/14} &  \textbf{65.45} &  \textbf{26.12} &  \textbf{39.07} &  \textbf{67.93} &  \textbf{29.27} &  \textbf{36.30} &  \textbf{73.98} &  \textbf{79.90} &  \textbf{29.51} &  \textbf{72.12} &  \textbf{60.20} &  \textbf{30.38} \\

\bottomrule
\end{tabular}
}
\label{tab:main_results}
\end{table*}

\begin{figure*}[ht]
    \centering
    \begin{subfigure}{.23\linewidth}
        \centering
        \includegraphics[width=\linewidth]{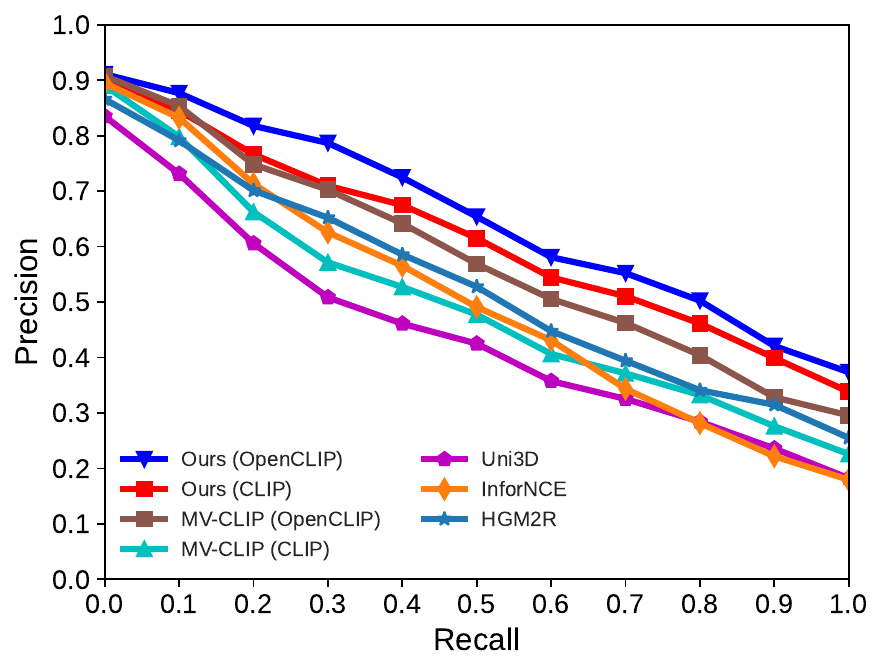}
        \caption{OS-ESB-core}
        \label{fig:esb}
    \end{subfigure}%
    \hfill
    \begin{subfigure}{.23\linewidth}
        \centering
        \includegraphics[width=\linewidth]{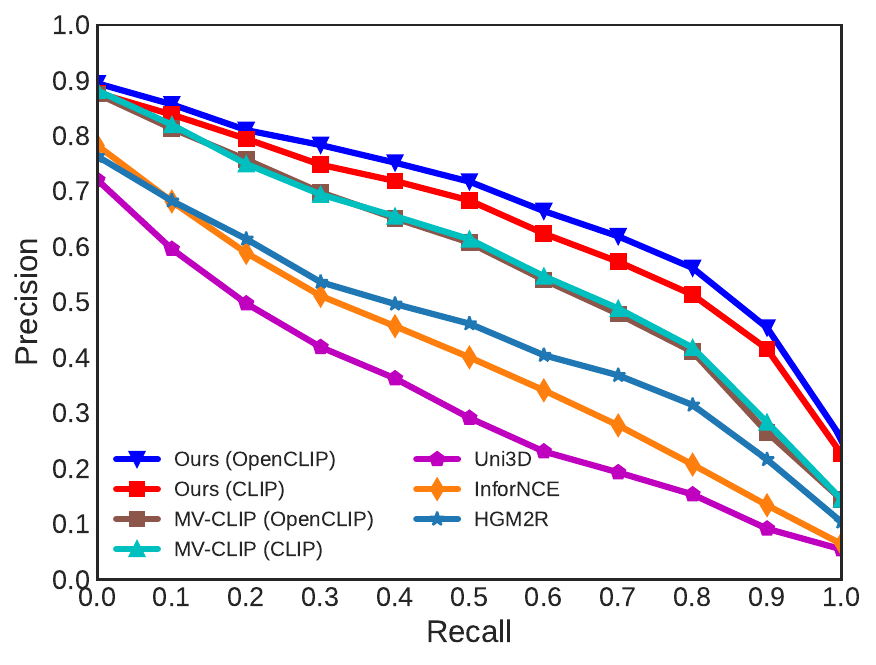}
        \caption{OS-NTU-core}
        \label{fig:ntu}
    \end{subfigure}%
    \hfill
    \begin{subfigure}{.23\linewidth}
        \centering
        \includegraphics[width=\linewidth]{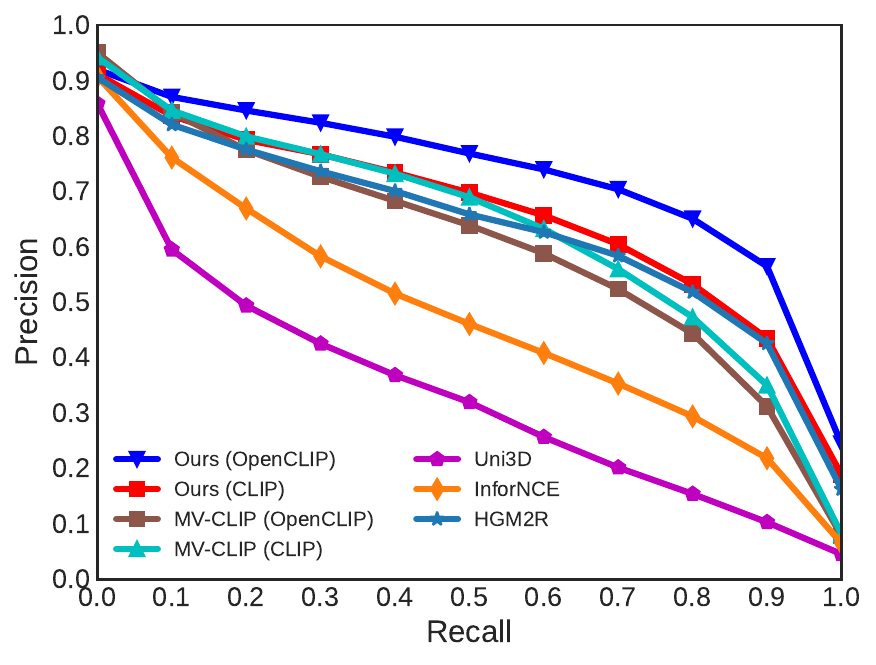}
        \caption{OS-MN40-core}
        \label{fig:mn40}
    \end{subfigure}%
    \hfill
    \begin{subfigure}{.23\linewidth}
        \centering
        \includegraphics[width=\linewidth]{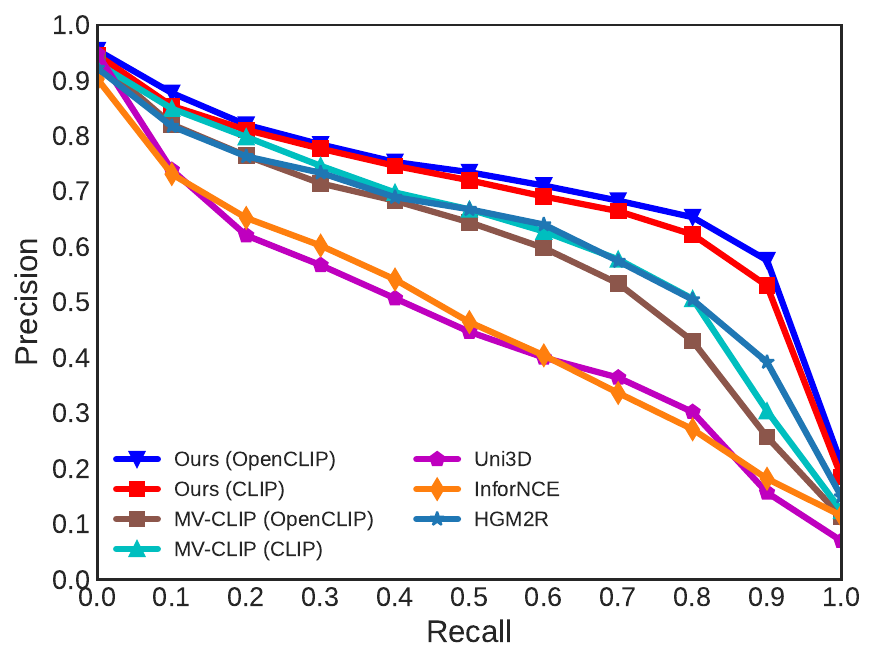}
        \caption{OS-ABO-core}
        \label{fig:abo}
    \end{subfigure}
    \caption{The precision-recall curve comparisons on the four datasets, respectively.}
    \label{fig:PRcurve}
\end{figure*}

\noindent\textbf{Compared Methods.}
To evaluate TeDA's effectiveness, we compare it against 13 representative methods, which can be roughly divided into four categories: 1) Vision-language-based 3D methods, \emph{i.e.}, OpenShape~\cite{liu2024openshape}, ULIP~\cite{xue2024ulip}, ULIP-2~\cite{xue2024ulip}, and Uni3D~\cite{zhou2023uni3d}, which improve 3D representation learning with the help of CLIP-based embeddings. 
We also re-implement MV-CLIP~\cite{song2023mv}, which is a strong zero-shot method \emph{but} requires a ground-truth label set for view selection. 
2) Existing superior 3DOR methods including TCL~\cite{he2018triplet},  
SDML~\cite{hu2019scalable}, and CMCL~\cite{jing2021cross}). They have demonstrated strong performance in the closed-set setting and are now extended to the new setup. 3) Auto-encoder-based methods (MMSAE~\cite{wu2019multi}, MCWSA~\cite{zheng2022multi}), which attempt to learn compressed representations in an unsupervised paradigm. 
4) Specially-designed open-set methods (PROSER~\cite{zhou2021learning}, InfoNCE~\cite{oord2018representation} and HGM$^{2}$R~\cite{feng2023hypergraph}), 
which is designed to deal with generalized representation learning of unknown categories. Except for Vision-language-based 3D methods, which operate in a zero-shot setting, all other methods require retraining on downstream datasets for optimal performance.

\noindent\textbf{Result Analysis.}
Table~\ref{tab:main_results} compares retrieval performance between our method and other representative methods.
We conduct experiments under two settings: \emph{Training-required} and \emph{Training-free}.
Our method significantly outperforms all others, both in the \emph{Training-required} and \emph{Training-free} settings. Specifically, compared to the state-of-the-art method HGM$^2$R in the \emph{Training-required} setting, using OpenCLIP ViT-B/14 as the backbone, we achieve improvements of 13.71\% and 23.05\% on the OS-ESB-core and OS-NTU-core datasets, respectively. Moreover, unlike HGM$^2$R, our method does not 1) require additional time for training and 2) rely solely on multi-view images, but instead leverages multi-modality (voxels, point clouds, multi-view images). 
In the \emph{Training-free} setup, TeDA also outperform existing approaches, such as Uni3D, which is pretrained on a large-scale 3D dataset, and MV-CLIP, a zero-shot method based on view selection. 
Overall, our method offers two key advantages: 1) It leverages a 2D pre-trained model and only adapts the extracted embeddings at test time, without the need for extensive 3D data for training; and 2) it requires no model fine-tuning and achieves remarkable results with only minimal time spent on extracting descriptions and optimizing retrieval distributions. 

Finally, for comprehensive comparisons, we further plot Precision-Recall (PR) Curves (Figure~\ref{fig:PRcurve}). As shown, our TeDA consistently yields better curves against all other opponents. 

\begin{table*}[h]
    \centering
    \small
    \caption{Effectiveness of the designed modules. Values are presented in mAP/NDCG/ANMRR format.}
    \setlength
    \tabcolsep{10pt}
\resizebox{0.95\textwidth}{!}{%
    \begin{tabular}{ccccccc}
        \toprule
        Backbone & InternVL & DA & OS-ESB-core & OS-NTU-core & OS-MN40-core & OS-ABO-core \\
        \midrule
        \multirow{4}{*}{CLIP ViT-L/14} 
        & \xmark & \xmark & 54.68 / 23.39 / 48.67 & 57.29 / 26.24 / 45.47 & 55.01 / 70.55 / 45.72 & 57.35 / 56.13 / 44.42 \\
        & \cmark & \xmark & 56.55 / 23.93 / 47.45 & 61.33 / 27.57 / 41.84 & 59.50 / 72.73 / 41.75 & 65.78 / 59.22 / 36.57 \\
        & \xmark & \cmark & 60.92 / 25.13 / 43.99 & 61.63 / 27.44 / 41.77 & 63.94 / 73.83 / 38.27 & 65.59 / 57.80 / 37.02 \\
        & \cmark & \cmark & \textbf{61.40} / \textbf{25.30} / \textbf{43.74} & \textbf{64.49} / \textbf{28.45} / \textbf{39.62} & \textbf{66.31} / \textbf{74.98} / \textbf{36.22} & \textbf{69.91} / \textbf{59.30} / \textbf{32.41} \\
        \midrule
        \multirow{4}{*}{OpenCLIP ViT-L/14} 
        & \xmark & \xmark & 62.06 / 25.11 / 42.09 & 62.84 / 27.78 / 39.78 & 64.97 / 76.55 / 36.99 & 61.29 / 57.06 / 40.90 \\
        & \cmark & \xmark & 62.18 / 25.22 / 41.91 & 66.08 / 28.76 / 37.78 & 68.10 / 77.98 / 34.23 & 67.88 / 59.77 / 34.49 \\
        & \xmark & \cmark & 65.05 / 26.06 / 40.24 & 66.47 / 28.91 / 37.74 & 73.42 / 79.65 / 29.94 & 69.97 / 58.66 / 32.36 \\
        & \cmark & \cmark & \textbf{65.45} / \textbf{26.12} / \textbf{39.07} & \textbf{67.93} / \textbf{29.27} / \textbf{36.30} & \textbf{73.98} / \textbf{79.90} / \textbf{29.51} & \textbf{72.12} / \textbf{60.20} / \textbf{30.38} \\
        \bottomrule
    \end{tabular}
    }
     \label{tab:ablation}
\end{table*}

\subsection{Analyses}
\noindent\textbf{Impact of InternVL.}
InternVL~\cite{chen2024internvl} plays a critical role in recognizing unknown categories. By inputting multi-view images of 3D objects and prompts into InternVL, it generates descriptive text for corresponding objects. As shown in Table~\ref{tab:ablation}, we observe consistent improvements after incorporating the textual cues. However, the magnitude of improvement varies significantly among datasets. For example, with CLIP ViT-L/14 as the backbone, the performance on OS-ESB-core improves by only 1.87\%, while on the OS-ABO-core dataset, it achieves an 8.43\% performance boost. This disparity is correlated with the dataset complexity. OS-ESB-core primarily contains high-genus objects~\cite{jayanti2006developing}, such as mechanical parts, whereas OS-ABO-core consists of common objects like chairs and beds.

\noindent\textbf{Impact of the Distribution Alignment Module.}
The Distribution Alignment (DA) Module filters high-confidence soft labels, which not only eliminates the influence of noisy labels but also enables the model to make high-confidence predictions for unknown categories. 
As shown in Table~\ref{tab:ablation}, applying DA results in significant performance improvements across all four datasets. For instance, when using CLIP ViT-L/14 as the backbone, the addition of DA (row 3) leads to remarkable improvements of 8.93\% and 8.24\% in mAP over the baseline (row 1) on OS-MN40-core and OS-ABO-core, respectively. 
The notable gains demonstrate the effectiveness of DA in bridging the distribution gaps at the test time.

\noindent\textbf{Design choices for InternVL Prompts.} We conducted ablation experiments on the prompts used in InternVL, selecting two representative prompts: Q1: ``There are images of an object from different angles. Describe this object in one sentence." and Q2: ``There are images of an object from different angles. Describe this object's shape information in one sentence." The experimental results, shown in Table~\ref{tab:prompt}, indicate that Q1 consistently performs better across all datasets. We hypothesize that more explicit descriptions help to enhance the discriminative power of the retrieval results. For example, "The object is a wooden table." is clearly more distinctive than "An object has a flat surface supported by four legs." However, InternVL may also make incorrect judgments about the category information, potentially affecting retrieval performance. Thus, the choice of prompt is well-worthy of future investigation.

\begin{table}[ht]
    \centering
    \small
    \caption{Impact of different prompts.}
    \resizebox{0.9\linewidth}{!}
    {
        \begin{tabular}{lcccc}
        \toprule
        Dataset & InternVL Prompt & mAP$\uparrow$ & NDCG$\uparrow$ & ANMRR$\downarrow$ \\
        \midrule
        \multirow{2}{*}{OS-ESB-core} 
            & Q1 & 65.45 & 26.12 & 39.07 \\
            & Q2 & 64.84 & 26.08 & 39.74 \\
        \cmidrule{1-5}
        \multirow{2}{*}{OS-NTU-core} 
            & Q1 & 67.93 & 29.27 & 36.30 \\
            & Q2 & 67.44 & 29.14 & 36.61 \\
        \cmidrule{1-5}
        \multirow{2}{*}{OS-MN40-core} 
            & Q1 & 73.98 & 79.90 & 29.51 \\
            & Q2 & 73.73 & 79.68 & 29.74 \\
        \cmidrule{1-5}
        \multirow{2}{*}{OS-ABO-core} 
            & Q1 & 72.12 & 60.20 & 30.38 \\
            & Q2 & 71.35 & 59.08 & 31.24 \\
        \bottomrule
        \end{tabular}
    }
    \label{tab:prompt}
\end{table}

\begin{figure}[h]
\centering
\includegraphics[width=0.9\linewidth]{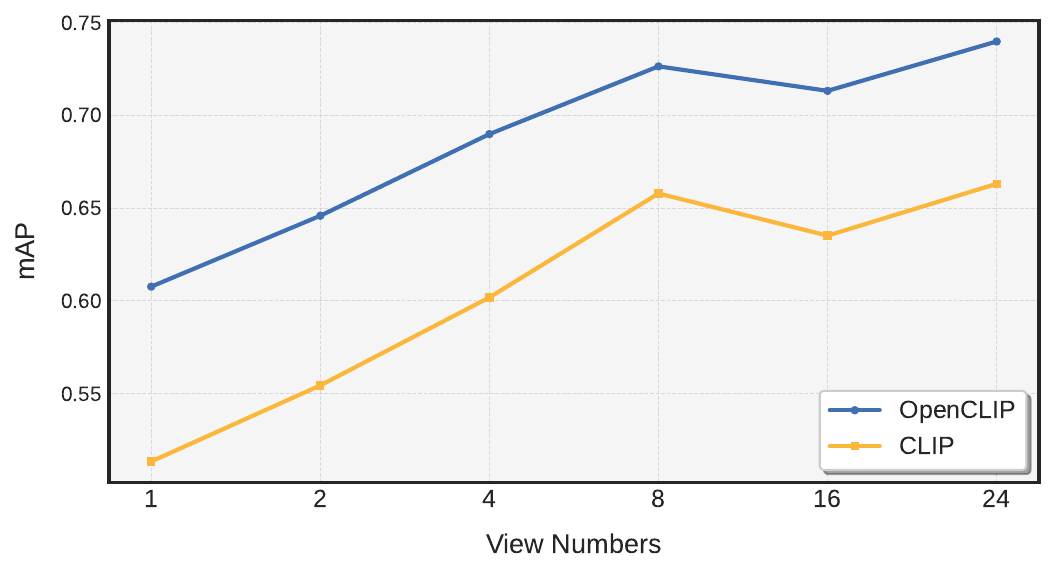}
\caption{Impact of View Numbers on OS-MN40-core.}
\label{fig:views}
\end{figure}

\noindent\textbf{Impact of View Numbers.} 
The number of views refers to both the number of projections in the multi-view setting and the number of views input into InternVL, which are identical in our experimental setup. As shown in Figure~\ref{fig:views}, performance generally improves with an increasing number of views. However, the result with 16 views is lower than that with 8 views, which is a consistent trend across both backbones. More views can provide a more comprehensive representation of the 3D object, but at the same time, they introduce redundant information that can negatively impact performance.

\begin{figure}[h]
\centering
\includegraphics[width=0.9\linewidth]{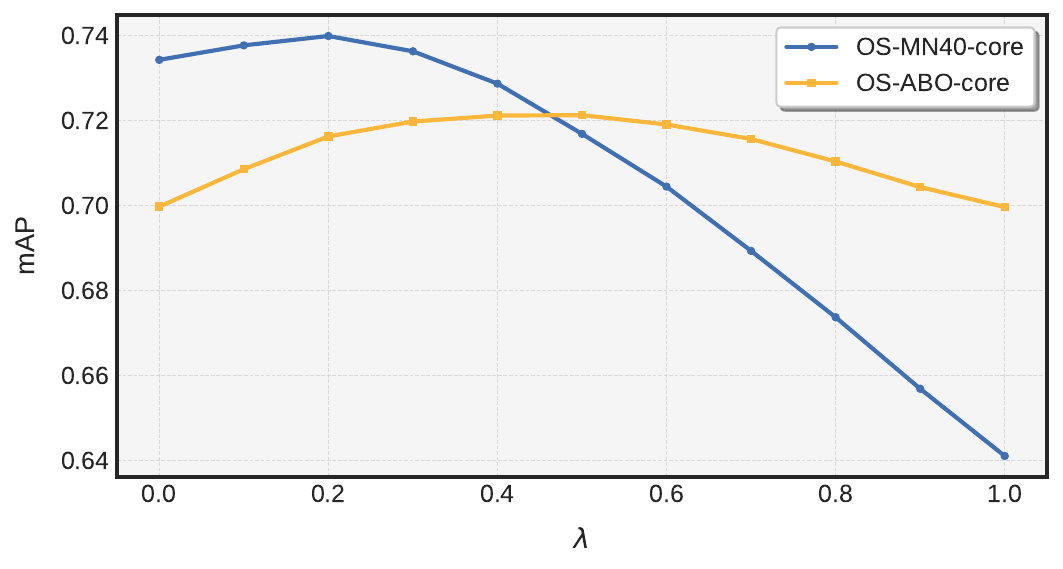}
\caption{Impact of Fusion Weight $\lambda$.}
\label{fig:fusion_weight}
\end{figure}

\noindent\textbf{Impact of Fusion Weight $\lambda$.}
The hyperparameter $\lambda$ controls the relative weight between text and image features. We adjust the fusion ratio $\lambda$ within the range of 0 to 1 and conduct experiments on OS-MN40-core and OS-ABO-core. As shown in Figure~\ref{fig:fusion_weight}, both datasets exhibit an increasing-then-decreasing trend with the rise of $\lambda$. Specifically, OS-MN40-core achieves the highest performance at $\lambda = 0.2$, while OS-ABO-core reaches its peak at $\lambda = 0.5$. We hypothesize that the value of $\lambda$ indirectly reflects the accuracy of the descriptions generated by InternVL. For OS-ABO-core, which primarily contains common objects such as chairs and beds, the generated descriptions are more precise, leading to optimal performance at higher values of $\lambda$. This hypothesis is further validated in Table~\ref{tab:ablation}, where on OS-ABO-core it shows a higher performance improvement with the application of InternVL.

\noindent\textbf{Impact of Threshold $\alpha$.} 
During the distribution alignment process, the threshold $\alpha$ is used to filter soft labels with high confidence, converting them into one-hot pseudo-labels. A suitable threshold strikes a balance between providing clear supervision signals and minimizing the influence of noisy instances. We conducted ablation experiments on the OS-ABO-core dataset to evaluate the impact of $\alpha$, as shown in Table~\ref{tab:ablation-alpha}. From the results, when the threshold is set to 1, the model directly uses the softmax-generated soft labels for distribution alignment. As the threshold decreases, more soft labels are converted into one-hot labels, which helps eliminate low-confidence noisy predictions and improves performance. On the other hand, when the threshold is reduced to 0, all soft labels are converted into one-hot encodings, amplifying the noise from low-confidence predictions and causing performance degradation. In all experiments of this paper, we selected $\alpha = 0.6$, which achieved the best performance, as the threshold. 

\begin{table}[ht]
    \centering
    \caption{Impact of threshold \( \alpha \).}
    \setlength{\tabcolsep}{10pt}
    \resizebox{0.9\linewidth}{!}{
    \begin{tabular}{lccccc}
    \toprule
    Backbone & Threshold \( \alpha \) & mAP$\uparrow$ & NDCG$\uparrow$ & ANMRR$\downarrow$ \\
    \midrule
        \multirow{5}{*}{OpenCLIP ViT-L/14} 
        & \textbf{\emph{0}} & 71.23 & 58.79 & 31.14 \\
        & \textbf{\emph{0.5}} & 71.82 & 59.81 & 30.46 \\
        & \textbf{\emph{0.6}} & \textbf{72.12} & \textbf{60.20} & \textbf{30.38} \\
        & \textbf{\emph{0.8}} & 71.79 & 60.07 & 30.61 \\
        & \textbf{\emph{1.0}} & 71.85 & 60.11 & 30.56 \\
    \bottomrule
    \end{tabular}}
\label{tab:ablation-alpha}
\end{table}

\begin{figure}[h]
\centering
\includegraphics[width=0.9\linewidth]{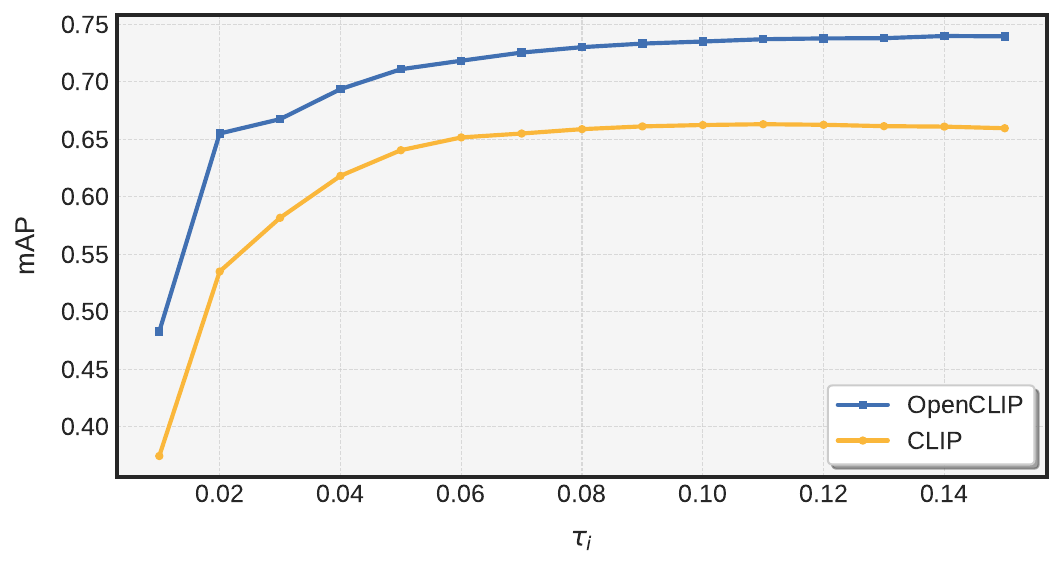}
\caption{Impact of Temperature $\tau_i$ with $\tau_t$ fixed at 0.03.}
\label{fig:temperature}
\end{figure}

\noindent\textbf{Impact of Temperature $\tau_i$.}
The temperature parameter controls the sharpness of the softmax output, where a higher temperature leads to a smoother distribution, and a lower temperature sharpens the distribution, making the model more confident in a few classes. We fixed $\tau_t$ at 0.03 and conducted an ablation experiment to investigate the impact of $\tau_i$ on OS-MN40-core, as shown in Figure~\ref{fig:temperature}. As shown in the figure, smaller $\tau_t$ and larger $\tau_i$ are more favorable for distribution alignment, as a smaller $\tau_t$ and a larger $\tau_i$ can calibrate the magnitude of the distribution. On the other hand, when $\tau_i$ is set to 0.01 (lower than $\tau_t$), the distribution optimization goal results in a highly sharpened output, leading to very poor performance.

\noindent\textbf{Impact of Fusion Schemes.}
We explored two parameter-free fusion methods: concatenation and element-wise addition. As shown in Table~\ref{tab:ablation-fusion}, element-wise addition achieves better results, which is beneficial because CLIP aligns the features of the two modalities during pretraining. 
\begin{table}[ht]
    \centering
    \caption{Impact of fusion Schemes on different backbones.}
    \setlength{\tabcolsep}{10pt}
    \resizebox{0.9\linewidth}{!}{
    \begin{tabular}{lccccc}
    \toprule
    Backbone & Fusion Method & mAP$\uparrow$ & NDCG$\uparrow$ & ANMRR$\downarrow$ \\
    \midrule
        \multirow{2}{*}{CLIP ViT-L/14} 
        & \textbf{\emph{Concat.}} & 54.23 & 63.08 & 47.24 \\
        & \textbf{\emph{Add.}} & \textbf{66.31} & \textbf{74.98} & \textbf{36.22} \\
        \midrule
        \multirow{2}{*}{OpenCLIP ViT-L/14} 
        & \textbf{\emph{Concat.}} & 62.05 & 70.33 & 40.31 \\
        & \textbf{\emph{Add.}} & \textbf{73.98} & \textbf{79.90} & \textbf{29.51} \\
    \bottomrule
    \end{tabular}}
\label{tab:ablation-fusion}
\end{table}

\noindent\textbf{Impact of Normalization.}
Normalizing the fused text and image features can further improve the generalization. We evaluate three commonly-used activation functions: \emph{ReLU}, \emph{Tanh}, and \emph{Sigmoid}. As shown in Table~\ref{tab:ablation-activation}. \emph{Tanh} yields the best performance, with a 2.59\% performance improvement when using CLIP ViT-L/14 as the backbone, while \emph{ReLU} and \emph{Sigmoid} perform worse than the baseline.
It can be attributed to the properties of the \emph{Tanh} function, which maps values to the range \([-1, 1]\), aligning well with CLIP's embedding space and producing evenly distributed outputs. In contrast, \emph{ReLU} eliminates negative values, potentially causing information loss, while \emph{Sigmoid} restricts outputs to \([0, 1]\), weakening vector directionality. Consequently, \emph{Tanh} is a more viable choice. 

\begin{table}[ht]
    \centering
    \small
    \caption{Impact of Activation Functions on OS-MN40-core.}
    \setlength{\tabcolsep}{10pt}
    \resizebox{0.9\linewidth}{!}{
    \begin{tabular}{lccccc}
    \toprule
    Backbone & Activation Function & mAP$\uparrow$ & NDCG$\uparrow$ & ANMRR$\downarrow$ \\
    \midrule
    \multirow{4}{*}{CLIP ViT-L/14} 
        & \textbf{-} & 63.72 & 72.32 & 38.71 \\
        & \textbf{\emph{ReLU}}  & 62.29 & 71.45 & 40.00 \\
        & \textbf{\emph{Sigmoid}}  & 58.46 & 70.81 & 43.05 \\
        & \textbf{\emph{Tanh}} & \textbf{66.31} & \textbf{74.98} & \textbf{36.22} \\
        \midrule
    \multirow{4}{*}{OpenCLIP ViT-L/14} 
        & \textbf{-} & 72.72 & 78.89 & 30.67 \\
        & \textbf{\emph{ReLU}}  & 72.41 & 78.86 & 31.01 \\
        & \textbf{\emph{Sigmoid}}  & 66.58 & 75.73 & 35.85 \\
        & \textbf{\emph{Tanh}} & \textbf{73.98} & \textbf{79.90} & \textbf{29.51} \\
    \bottomrule
    \end{tabular}}
\label{tab:ablation-activation}
\end{table}

\subsection{Extended Experiments on Depth Maps}

Our framework can also be effectively extended flexibly to point clouds by projecting it to 10 depth maps online~\cite{zhu2023pointclip}. 
We use Objaverse-LVIS dataset for zero-shot experiments, which is an annotated subset of Objaverse~\cite{deitke2023objaverse}, containing 46,832 shapes across 1,156 LVIS categories. We further split each category of Objaverse-LVIS into a query set and a target set with a 20\%/80\% ratio, resulting in a total of 8,798 query samples and 37,407 target samples. 
As shown in Table~\ref{tab:objaverse}, our method based on depth maps also achieves superior performance, surpassing ULIP-2 by 2.20\% in the mAP.

\begin{table}[ht]
    \centering
    \small 
    \caption{Performance on ZS-Objaverse-Core (Zero-shot).}
    \setlength{\tabcolsep}{10pt}
    \resizebox{0.9\linewidth}{!}{
    \begin{tabular}{lccccc}
    \toprule
    Method & Backbone & mAP$\uparrow$ & NDCG$\uparrow$ & ANMRR$\downarrow$ \\
    \midrule
    OpenShape (point cloud) & PointBERT-CLIP ViT-L/14 & 11.93\% & 14.10\% & 85.40\% \\
    ULIP (point cloud) & PointMLP - SLIP   & 6.69\% & 9.17\% & 90.82\% \\ 
    ULIP-2 (point cloud) & PointBERT - CLIP ViT-G/14  & 18.15\% & 19.34\% & 79.03\% \\ 
    Ours (depth images) & OpenCLIP ViT-L/14  & \textbf{20.35\%} & \textbf{20.40\%} & \textbf{77.48\%} \\ 
    \bottomrule
    \end{tabular}}
\label{tab:objaverse}
\end{table}

\subsection{Visualizations of Retrieval Examples}

\begin{figure}[h]
\centering
\includegraphics[width=0.9\linewidth]{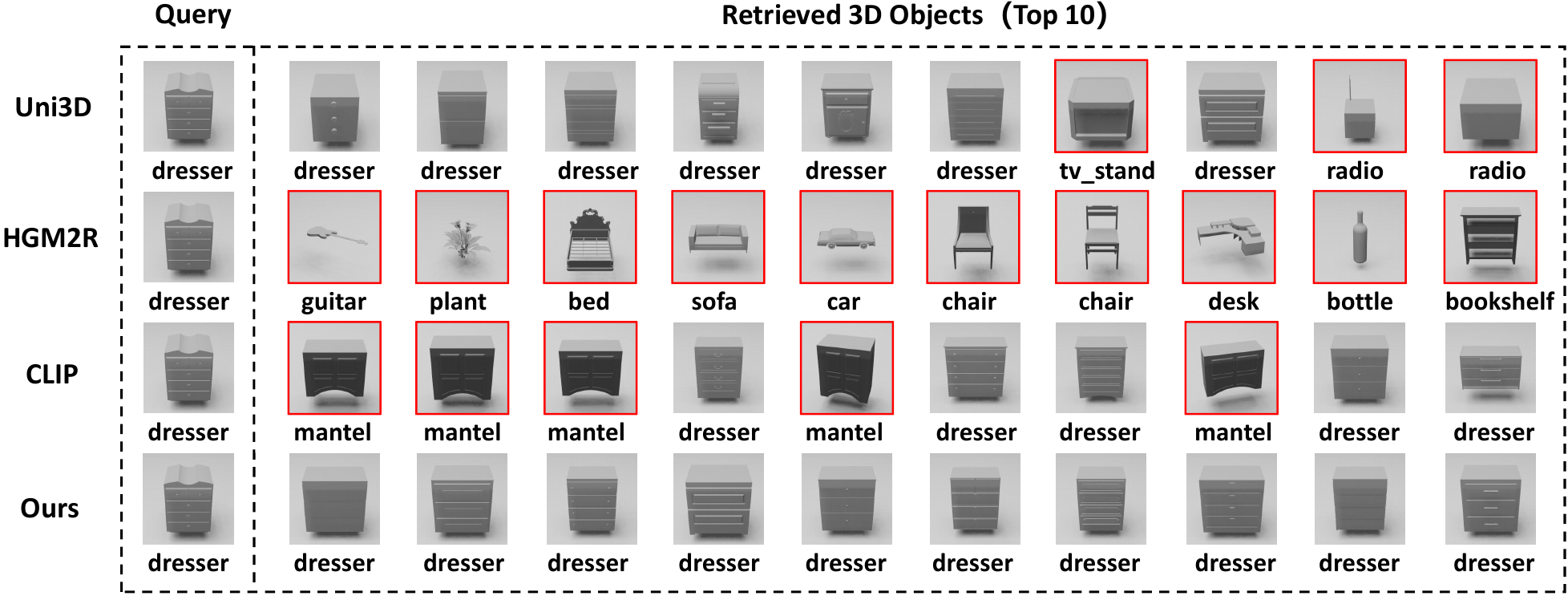}
\caption{Comparison of Retrieval Examples on OS-MN40-core. Incorrect matches are in red boxes. CLIP represents using CLIP's visual encoder to extract multi-view features and perform mean pooling for retrieval.}
\label{fig:vis}
\end{figure}

Figure~\ref{fig:vis} presents retrieval example comparisons with representative competitors. By incorporating output from InternVL, TeDA demonstrates superior fine-grained recognition, allowing it to capture subtle visual and semantic differences in unseen categories. For instance, CLIP mistakenly retrieves a mantel as a dresser due to similar visual features, whereas our method avoids such errors.

\section{Conclusion}

In this paper, we have proposed a novel framework named TeDA for unknown 3D representation learning, based only on a collection of multi-view images. 
TeDA only leverages a pretrained 2D vision-language model CLIP and adapts its extracted view embeddings \emph{at test time} with a novel test distribution alignment method in an unsupervised manner.  
Without requiring any 3D training dataset, our approach achieves significantly superior performance compared to both existing training-free and training-required methods. Furthermore, by incorporating textual modalities, our method further enhances retrieval performance, benefiting from the strong generalization capabilities of language representations, which is particularly crucial for retrieving unknown object classes. Additionally, we have extended our approach to depth maps, further validating its effectiveness with excellent results.

\section{Acknowledgments}
This work is supported by National Natural Science Foundation of China (No.62302188); Hubei Province Natural Science Foundation (No.2023AFB267); Fundamental Research Funds for the Central Universities (No.2662023XXQD001).

\bibliographystyle{ACM-Reference-Format}
\bibliography{ref}
  
\end{document}